\documentclass[preprint, 10pt]{elsarticle}
\pdfoutput=1
\usepackage{graphicx}
\usepackage{subcaption}
\usepackage{amssymb}
\usepackage{float}

\begin{document}
\begin{frontmatter}
\title{ActivationNet: Representation learning to predict contact quality of interacting 3-D surfaces in engineering designs}
\author{Rishikesh Ranade\corref{cor1}\fnref{label1}}
\cortext[cor1]{Corresponding Author.: Email Address: rishikesh.ranade@ansys.com (R.Ranade)}

\author[label1]{Jay Pathak}
\address[label1]{Ansys Inc., Canonsburg, Pennsylvania, USA}

\begin{abstract}
Engineering simulations for analysis of structural and fluid systems require information of contacts between various 3-D surfaces of the geometry to accurately model the physics between them. In machine learning applications, 3-D surfaces are most suitably represented with point clouds or meshes and learning representations of interacting geometries form point-based representations is challenging. The objective of this work is to introduce a machine learning algorithm, \textit{ActivationNet}, that can learn from point clouds or meshes of interacting 3-D surfaces and predict the quality of contact between these surfaces. The \textit{ActivationNet} generates activation states from point-based representation of surfaces using a multi-dimensional binning approach. The activation states are further used to contact quality between surfaces using deep neural networks. The performance of our model is demonstrated using several experiments, including tests on interacting surfaces extracted from engineering geometries. In all the experiments presented in this paper, the contact quality predictions of \textit{ActivationNet} agree well with the expectations.  
\end{abstract}



\end{frontmatter}

\section{Introduction} \label{introduction}

Many applications in engineering simulation \& robotics involve geometries with a large number of 3-D surfaces and a non-trivial interaction between them. The goal of engineering simulation is to predict the underlying physics present in a physical system or process. In many cases, the accuracy of these predictions is predicated on the contact between the various 3-D surfaces in the geometry. As a result, it is important to identify the contact quality of all the interacting 3-D surfaces in a geometry so as to effectively model the physical phenomenon resulting from their interactions.

Engineering simulation software, such as Ansys, use contact detection algorithms \cite{heinstein1993general, munjiza1995combined, munjiza1999combined, rougier2009discrete, munjiza2011computational, schiava2013novel} to identify all possible contacting surfaces in a given 3-D geometry. Due to algorithmic limitations, most of these methods only identify the interacting surfaces in the geometry but cannot separate surfaces with good contacts from bad. Surfaces with bad contact are undesirable and generally hamper the accuracy and convergence of physics-based solvers. Hence, it is necessary to prune these contacts before running the simulations. 

In engineering applications, the contact between 3-D surfaces is generally characterized by handcrafted features such as, proximity, solid angle and overlap. Proximity refers to the distance between the interacting surfaces, solid angle is the angle between them and overlap denotes the intersection between these surfaces. In general, 3-D surfaces with good contact should have a smaller proximity and angle but a larger overlap. Although, the computation of these features is extremely expensive and in some cases not possible at all, especially for arbitrarily shaped 3-D surfaces with curvy and intricate geometries. Additionally, each side of the contacting surfaces may be a collection of many surfaces which further adds to complexity of defining a good contact. Moreover, even if these features could be evaluated, it is very difficult to devise a metric for a contact quality score that includes a combination of these features, since the correlation between them may be non-linear depending on the size and shape of the 3-D surfaces. As a result, expert design modelers need to intervene and manually separate surfaces with good contacts from bad, solely based on experience and prior knowledge. In this context, machine learning strategies can be very useful in informing the design modelers on the nature and quality of the various contacting surfaces present in 3-D geometries, and save computational time and resources.

In recent times, Machine Learning (ML) algorithms have made significant progress in learning and representing complicated 3-D surfaces. The initial efforts in learning from 3-D geometries were made by developing and employing 3-D Convolutional Neural Networks (CNNs) \cite{maturana2015voxnet, yi2016scalable, qi2016volumetric}. These approaches worked with voxelized representations of geometries. As a result, they were prone to data sparsity and memory constraints. Research works following this tackled these challenges by representing 3-D surfaces and parts with point clouds and meshes. In the context of engineering computer aided design (CAD), 3-D geometries are generally represented using standard triangle language (STL). Point clouds represent the $x, y, z$ coordinates of nodes of the triangles representing the surfaces. On the other hand, meshes are represented by coordinates of mesh nodes as well as the adjacency relationship between them. The main advantage of point clouds and meshes is in their unstructured data structure, which can effectively represent the most intricate features of the 3-D geometry. However, feature extraction from these unstructured representations has proven to be challenging for deep learning algorithms. As a result, traditional approaches to feature extraction from point clouds relied on hand-crafting these features \cite{grilli2017review, johnson1999using, chen20073d, zhong2009shape, rusu2008aligning, rusu2009fast, tombari2010unique, chen2003visual, hansch2014comparison, ling2007shape}. The extracted features encode statistical information and are designed to be invariant to certain transformations. In recent years, other approaches of converting point clouds and meshes to voxel-based structured representations have been explored \cite{zhou2018voxelnet, maturana20153d, wang2019normalnet, ghadai2018multi, wu20153d}. The voxel-based representations allow feature extraction using the already matured CNN-based network architectures, but these approaches are memory intensive, especially for 3-D data sets. Other alternative approaches include using multi-view image representations of point cloud \cite{su2015multi, leng20153d, bai2016gift, kalogerakis20173d, cao20173d, zhang20183d}. Multi-view CNNs provide a better performance than voxel-based approaches but lose accuracy in the stitching process. More recently, CNNs have also been extended with multi-resolution voxel representation to efficiently handle memory requirements in encoding and decoding 3-D geometries \cite{wang2017cnn, wang2018adaptive}. The advent of new modeling approaches such as PointNet \cite{qi2017pointnet}, PointNet++ \cite{qi2017pointnet++} and graph CNNs \cite{klokov2017escape, wang2019dynamic, wang2018local, zhang2019shellnet, han2019point2node} provide an opportunity to learn features directly from point cloud and mesh representations and have applications in tasks involving classification, segmentation and detection. However, to the knowledge of the authors, the research work using these methods has focused on extracting features from geometries of single 3-D parts, and has never been applied to evaluate the interactions between multiple contacting 3-D surfaces.

In this work, we introduce the \textit{ActivationNet}, which is a machine learning algorithm that uses coarse point cloud or mesh representations of interacting 3-D surfaces and predicts a contact quality score associated to the surface pair. The \textit{ActivationNet} uses voxel-based methodologies to pre-process the interacting 3-D surface points and compute activation states using a multi-dimensional binning strategy \cite{van2014scikit}. The activation states are based on the spatial location and density of points, represented either by point clouds or mesh nodes, in a normalized 3-D space. The activation states provides neural networks the opportunity to extract features which are equivalent to the hand-crafted features such as proximity, angle and overlap. It also takes into account the important and intricate characteristics of 3-D surfaces, that are useful for learning surface interactions. A detailed explanation of the meaning and construction of activation states as well as a description of the network architecture of \textit{ActivationNet} has been provided in Section \ref{ActivationNet} of this paper.  

The rest of the paper is organized as follows. In section \ref{ActivationNet}, we describe the \textit{ActivationNet} in great detail and describe the generation of activation states. This is followed by a description of the data generation process and the training mechanics. The \textit{ActivationNet} is validated on several interacting 3-D surfaces extracted from unseen 3-D geometry models of engineering objects. The results from the \textit{ActivationNet} are compared with those from PointNet to showcase the ability of the \textit{ActivationNet} to learn representations and extract features from point clouds of 3-D surfaces, to predict contact quality with a very high accuracy.

The key contributions of our work are as follows:

\begin{itemize}
    \item We propose a machine learning methodology, \textit{ActivationNet} to extract features from interacting point clouds to predict a contact quality score.
    \item We demonstrate that \textit{ActivationNet} learns true representations which are useful in making intuitive predictions and it is not just overfitting data.
    \item Furthermore, our approach is demonstrated for predicting contact quality on actual engineering geometries used in engineering simulations pertaining to structural analysis.
\end{itemize}

\section{\textit{ActivationNet}: Model formulation} \label{ActivationNet}

\textit{ActivationNet} is a machine learning algorithm that predicts contact quality from point-based representations of interacting 3-D surfaces. The point-based representation is an unavoidable starting point for our algorithm because all engineering simulations use either point clouds or meshes to represent 3-D geometries. As stated previously, representation learning from discrete points is challenging, and hence we extract features not directly from these points but from the activation states, which are derived from them.

\begin{figure}
\begin{center}
\includegraphics[width=1.0\linewidth]{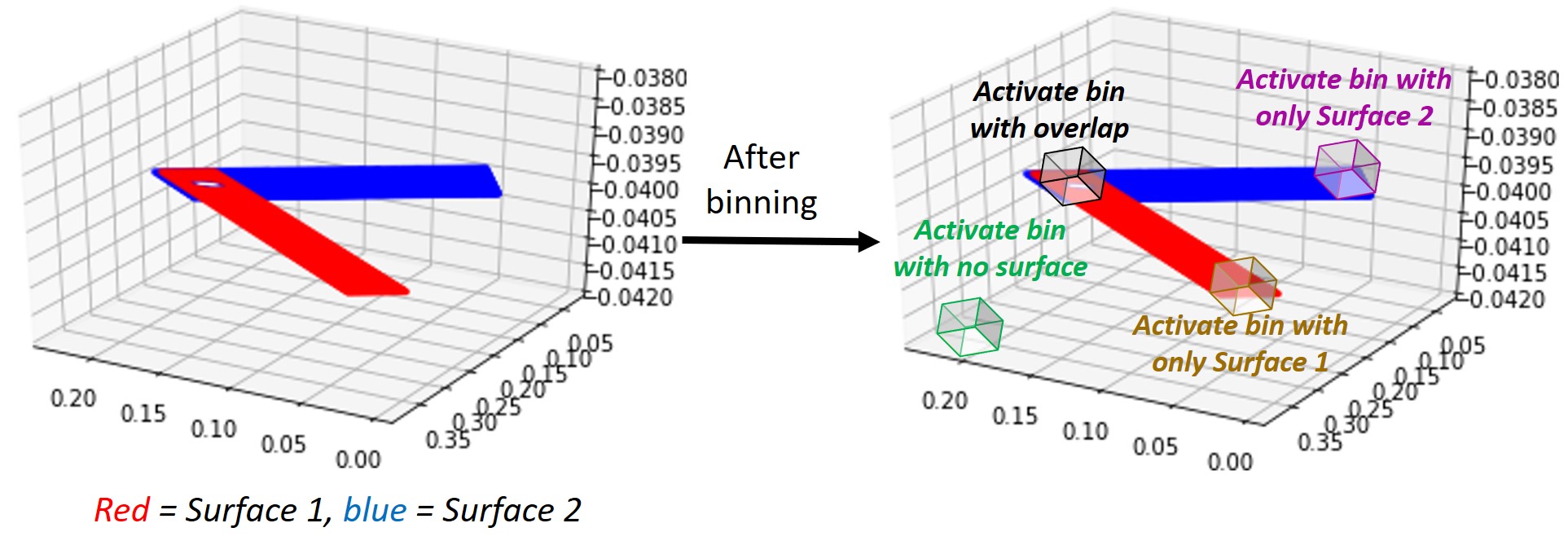} 
\end{center}
   \caption{Activation state using multi-dimensional binning}
\label{fig:fig1}
\end{figure}

Activation states are constructed using the multi-dimensional binning approach. Starting from discrete points representing interacting surfaces, the binning approach divides the 3-dimensional space occupied by these points into smaller bins of uniform size and shapes. This process is similar to voxelization except that, each bin is characterized by the number of points it contains and information regarding the interacting surface they belong to. These bin characteristics are used to determine an activation state for each bin. Figure \ref{fig:fig1} describes the construction of activation states in more detail. In Figure \ref{fig:fig1}, the surface $1$ is represented by a dense cloud of points marked red, while the surface $2$ is represented by points marked blue. The multi-dimensional binning approach divides the space surrounding the point clouds into smaller cube boxes. The bounds of the surrounding 3-D space are determined based on the span of the coordinates of the points. As mentioned earlier, each bin is associated an activation state based on the characteristics of the bin. For example, a bin containing points from only surface $1$ is given an activation state, say $1$, while a bin with points from only surface $2$ is given an activation state, say $2$. On the other hand, bins with points from both surfaces are considered to be bins representing overlapping regions of the interacting surfaces, and are given a different activation state, say $3$. Finally, bins with points from neither surfaces are simply activated with $0$. Moreover, this can be easily extended to scenarios with more than $2$ interacting surfaces. In such cases, the binning procedure would remain the same, but the activation states would also take into account the interaction between all the pairs of different surfaces. To that end, the activation states on each bin implicitly represent the extent of overlapping regions, based on the number of overlapping bins for a given sample, proximity and angle between the surfaces, based on the proximity and orientation of a given bin with respect to other bins. Thus, activation states provide neural networks the opportunity to directly extract these features and estimate contact quality.

\begin{figure}[h]
\begin{center}
\includegraphics[width=0.8\linewidth]{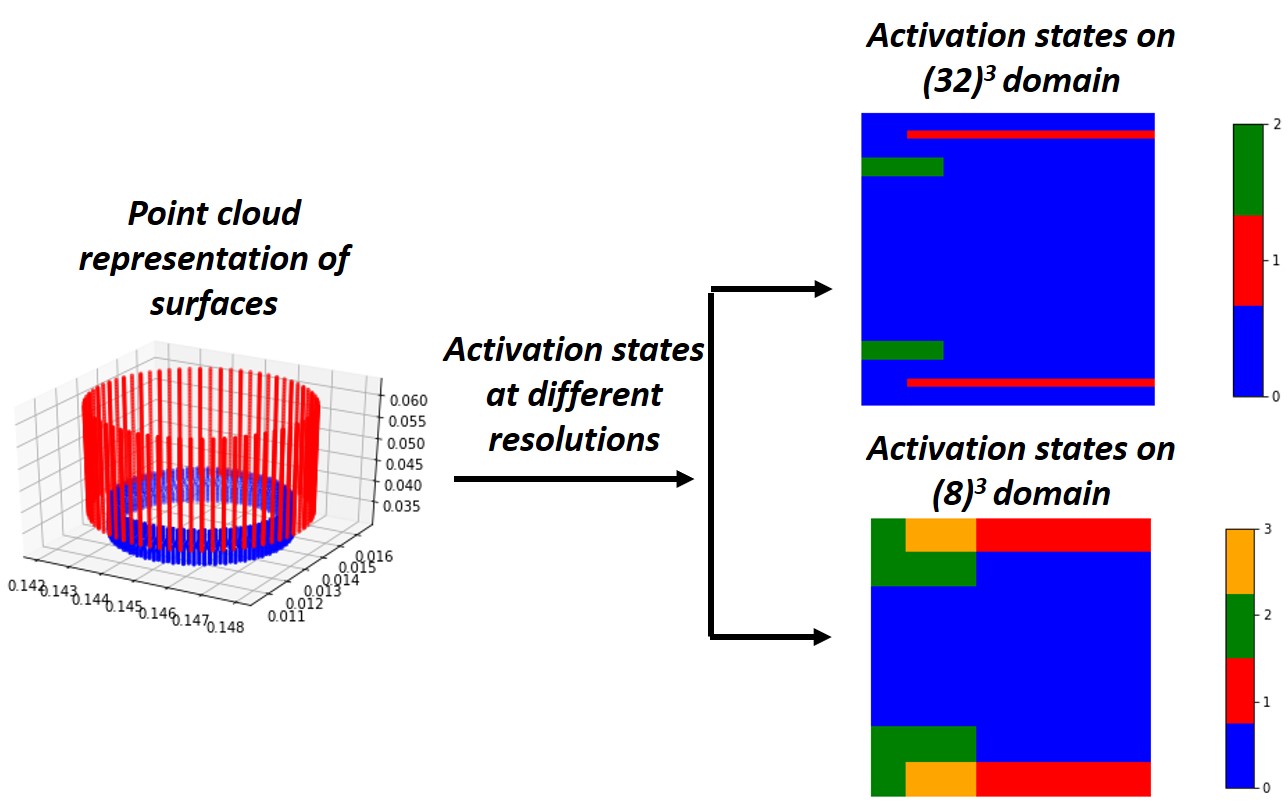} 
\end{center}
   \caption{Multi-resolution activation states}
\label{fig:fig2}
\end{figure}

\begin{figure*}[h]
\begin{center}
\includegraphics[width=1.0\linewidth]{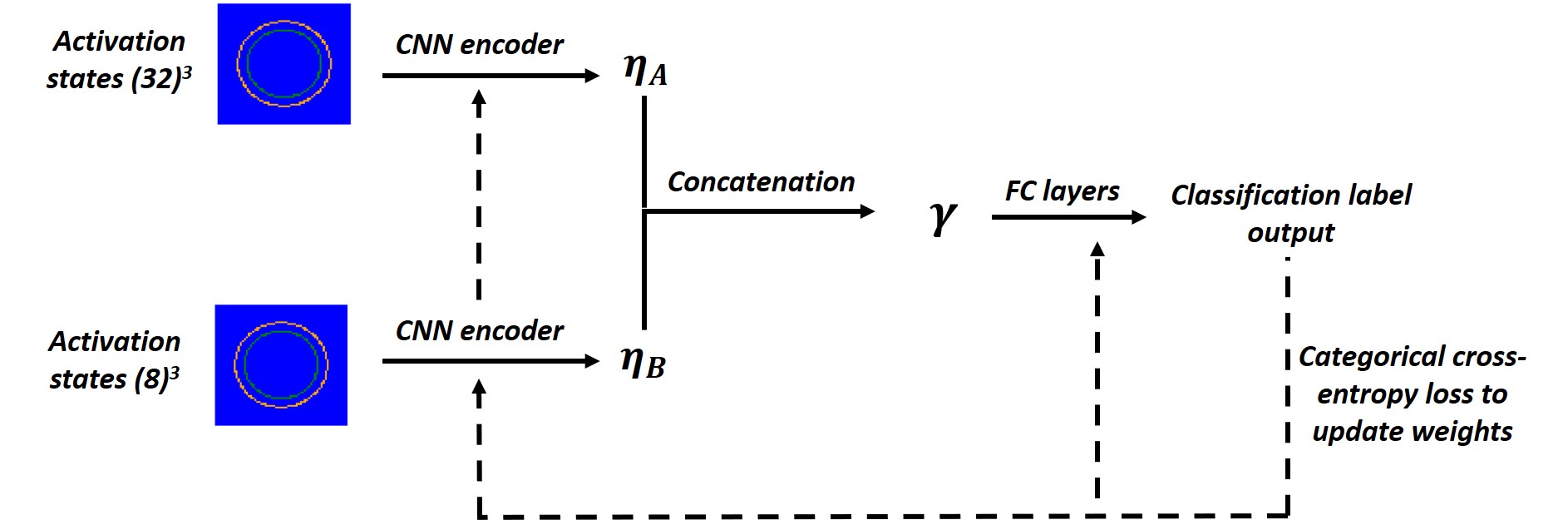} 
\end{center}
   \caption{Neural network architecture}
\label{fig:fig3}
\end{figure*}

It may be important to note that the features represented by activation states depends on the choice of bin size in the multi-dimensional binning approach. Larger bin sizes will extract coarser features of surface interaction as opposed to smaller bins, which will focus more on the finer and intricate features. Features from both representations may be necessary to accurately predict the quality of surface interactions. Figure \ref{fig:fig2} shows a 2-D cut plane drawn at an intercept of $0.04$ in the vertical direction. It may be observed that the lower resolution bins have a coarser representation, while the finer resolution bins have a much crisper representation. Both coarser and finer representations are important to understand the relative position of the interacting surfaces with each other. For example, in Figure \ref{fig:fig2}, the partial overlap in the coarser resolution indicates that the interacting surfaces may have a projected overlap but the finer resolution does not capture this feature but focuses more on the distance between the two surfaces. In many cases, the interplay between the projected overlap and proximity can influence the overall quality of the contacting surfaces.  

Thus, the activation states at multiple resolutions are used as inputs in the neural network. Figure \ref{fig:fig3} shows a schematic diagram of the neural network architecture used in this work. First, the multi-resolution activation states are encoded using convolutional layers. The network uses a combination of convolutional layers followed by pooling layers to extract features at different spatial resolutions. The final convolutional layer at the smallest spatial resolution is flattened to determine an encoding for each activation states. The encoding of the higher resolution activation state is represented by $\eta_a$, while the lower resolution activation state is represented by $\eta_b$. The size of the encoding vectors may be different for different resolutions. The encoding of activation states may capture a combination of features such as proximity, angle and overlap but are not limited to that. The activation states' encodings are concatenated and passed through fully connected deep layers. Both convolutional and fully connected layers use a ReLU activation function. Moreover, all the convolutional layers are followed by batch normalization layers \cite{ioffe2015batch} and the fully connected layers by dropout layers \cite{srivastava2014dropout}. The output layer of the network predicts the class probability and hence has a sigmoid activation. This problem is setup as a three class problem, where the class label $1$ refers to good contacts (with a quality score of $100$), class label $2$ refers to bad contacts (with a quality score of $0$) and class label $3$ refers to neutral contacts (with a quality score of $50$), which are neither good nor bad and have a quality of somewhere in between. Finally, a heuristic computation of contact quality score is designed based on the class probabilities predicted by the neural network. The contact quality score for a given pair of interacting surfaces is shown below in equation \ref{eq:eq1}.
\begin{equation}\label{eq:eq1}
    C = P(1)*100.0 + P(2)*0.0 + P(3)*50.0
\end{equation}
where, $C$ is the contact quality score and $P$ is the class probability. 

A three class problem was found to be very suitable for this application because the contact quality calculation is based on the class probabilities of network output. A two-class problem might bias the contact quality prediction in one direction or the other. The third class puts the quality score on a spectrum and accommodates for interacting 3-D surfaces whose contact quality is neither good or bad and depends largely on the application.

Since this problem is set up as a three-class classification problem, we use a categorical cross-entropy loss to update the network weights. The training is carried out using TensorFlow using a mini-batch Adams optimizer. The training is started with a learning rate of $1e^-3$ and a learning rate scheduler is used to modulate the learning rate depending on the validation losses. In Section, \ref{datagen}, we discuss the data generation and annotation in more detail.

\section{Data generation and annotation} \label{datagen}

\begin{figure*}[h]
\begin{center}
\includegraphics[width=1.0\linewidth]{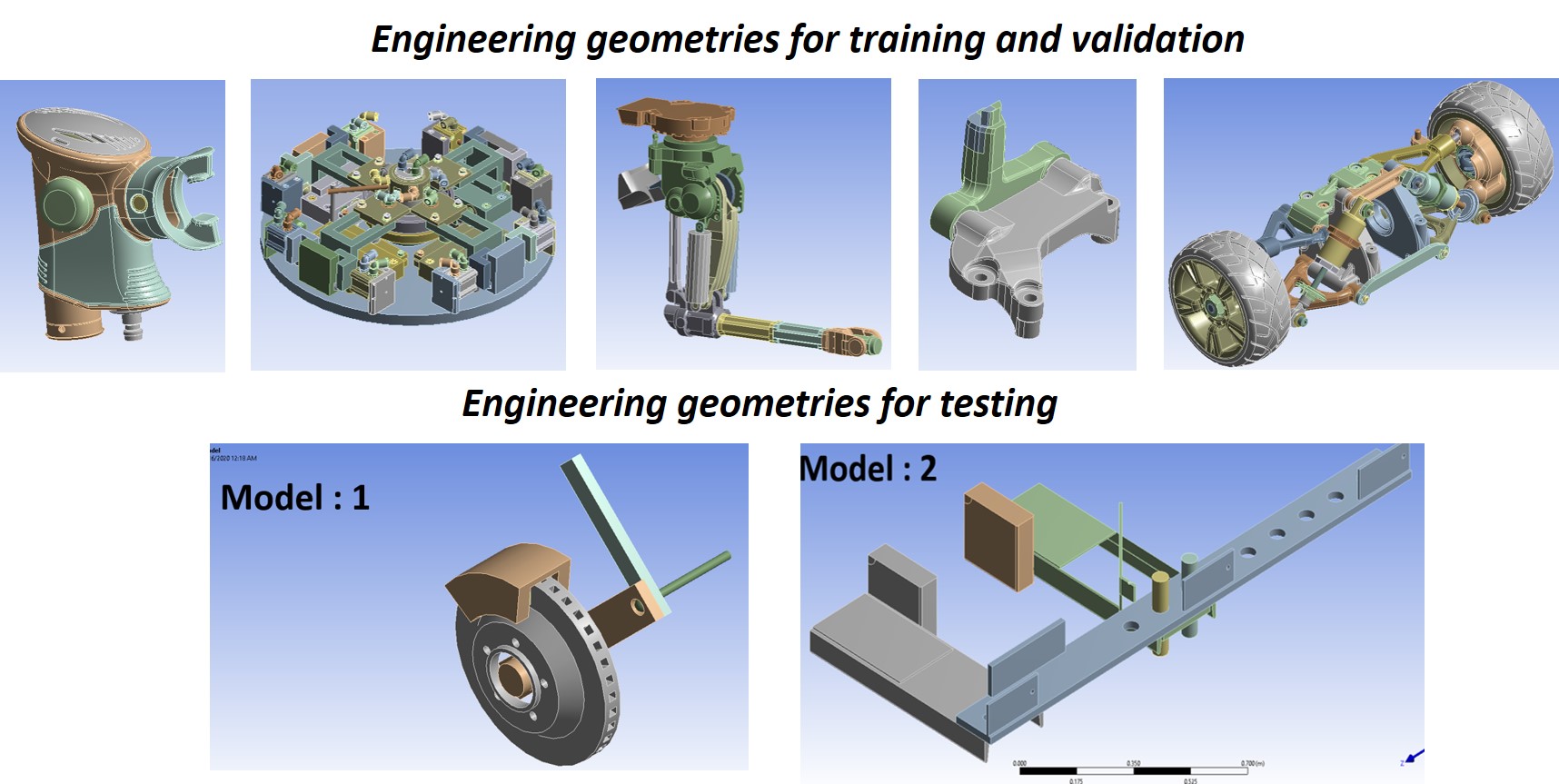} 
\end{center}
   \caption{Engineering geometries used for creating training, validation and testing datasets}
\label{fig:fig4}
\end{figure*}

The training and validation data for our network consists of 3-D interacting surfaces extracted from $5$ CAD models of engineering geometries, while the testing data consists of surfaces extracted from $2$ different CAD models. We use the Ansys software suite to model the engineering geometries. Figure \ref{fig:fig4} shows the engineering geometries used for generating the data. On each engineering geometry, we use a contact detection algorithm to identify all possible contacting 3-D surfaces. The algorithm uses a proximity-based single gap tolerance to identify surfaces that will be considered to be in contact. The gap tolerance itself is heuristically determined using a fraction of largest diagonal of all bodies in the CAD model. Since the contact detectors are not good at identifying contacts with good quality, ML algorithms such as the one proposed in this work, \textit{ActivationNet}, are extremely useful in accurately predicting contact quality, such that contacts with bad quality can be easily eliminated. 

Since, there is no explicit calculation for determining the contact quality score for arbitrary 3-D interacting surfaces, the training, validation and testing data is annotated by assigning a score range (between $0$ and $100$) based on visual inspection with guidance from experts in engineering simulation. The range of quality score assigned by expert modelers varies within $10$ points. In this work, the score ranges for each pair of 3-D surfaces are averaged over evaluations from $10$ expert design modelers. The averaged score range is used to assign a class label for all 3-D interaction surfaces. We assign samples to, 
\begin{itemize}
    \item class $1$, if score $>= 80$
    \item class $2$, if score $<= 20$
    \item class $3$, if score $>= 20$ and  $<=80$
\end{itemize}

The 3-D interacting surfaces are represented by discrete points, which are generated meshing the surfaces with coarse triangular meshes and extracting the nodal coordinate information. Sampling techniques may be used to generate more points within each triangular element if a higher resolution is required. The nodes of the triangular elements as well as any sampled points are used as point-based representations and are provided as inputs to the \textit{ActivationNet} algorithm described in Section \ref{ActivationNet}. The point representations of 3-D surfaces extracted from $5$ engineering geometries and their associated class labels are split into training and validation datasets, such that $75$\% is reserved for training and the rest for validation. There are only $300$ contacting surfaces extracted from the $5$ engineering geometries. Data augmentation techniques to rotate point clouds along planes in $x, y, z$ directions are used to increase the training data. On the other hand, as mentioned previously, 3-D surfaces extracted from completely new engineering geometries, which our model has never seen before, are used for testing purposes.

\section{Results \& discussions} \label{results}

In this section, we demonstrate the trained \textit{ActivationNet} for several cases and scenarios. In the first set of experiments, the goal is analyze whether the \textit{ActivationNet} is truly learning surface interactions and not just overfitting the training data set. The training data set consists of only $300$ arbitrarily shaped contacting surfaces from $5$ engineering geometries and hence, this experiment is crucial to prove the representational learning capability of our method. To that end, in this experiment we conduct demonstrations on simpler, interacting 3-D surfaces. These surfaces are translated, rotated and scaled relative to one another and the predictions obtained from the \textit{ActivationNet} are evaluated to prove that our approach is learning valid representations. In the next set of experiments, we demonstrate the \textit{ActivationNet} on 3-D surfaces extracted from the $2$ engineering geometries, that are reserved for testing. In this experiment, we also compare our results with the PointNet \cite{qi2017pointnet} algorithm for computation using point clouds.

\subsection{Translation of point clouds of 3-D surfaces}

\begin{figure}[h]
\begin{center}
\includegraphics[width=1.0\linewidth]{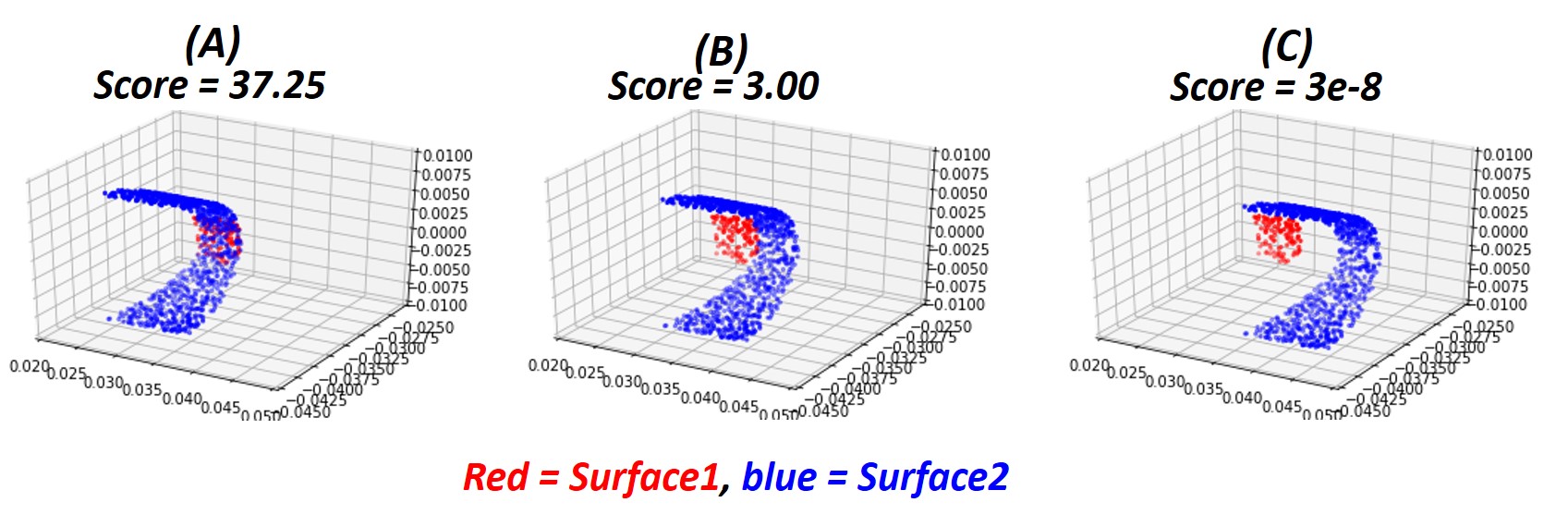} 
\end{center}
   \caption{\textit{ActivationNet} predictions for translation of 3-D surfaces}
\label{fig:fig5}
\end{figure}

In this study, we evaluate the predictions of \textit{ActivationNet} for translation of 3-D surfaces with each other. Intuitively, as 3-D surfaces move away from each other, the contact between them reduces. The goal of this study is to analyze the consistency of the \textit{ActivationNet} with respect to this observation. In Figure \ref{fig:fig5}, we show two surfaces which are initially in contact with each other and subsequently are pulled apart. The corresponding contact quality score predicted by our model is also shown in Figure \ref{fig:fig5}. It may be observed that initially when the surfaces are in contact with each other, the predicted quality score is $17.22$. The predicted contact score is relatively low, because, even though the surfaces are close to each other, the overlapping regions are small. Nonetheless, as the surfaces move further away from each other, the contact score predictions significantly drop and almost reach close to $0$ in the most extreme case.

\subsection{Rotation of point clouds of 3-D surfaces}

\begin{figure}[h]
\begin{center}
\includegraphics[width=1.0\linewidth]{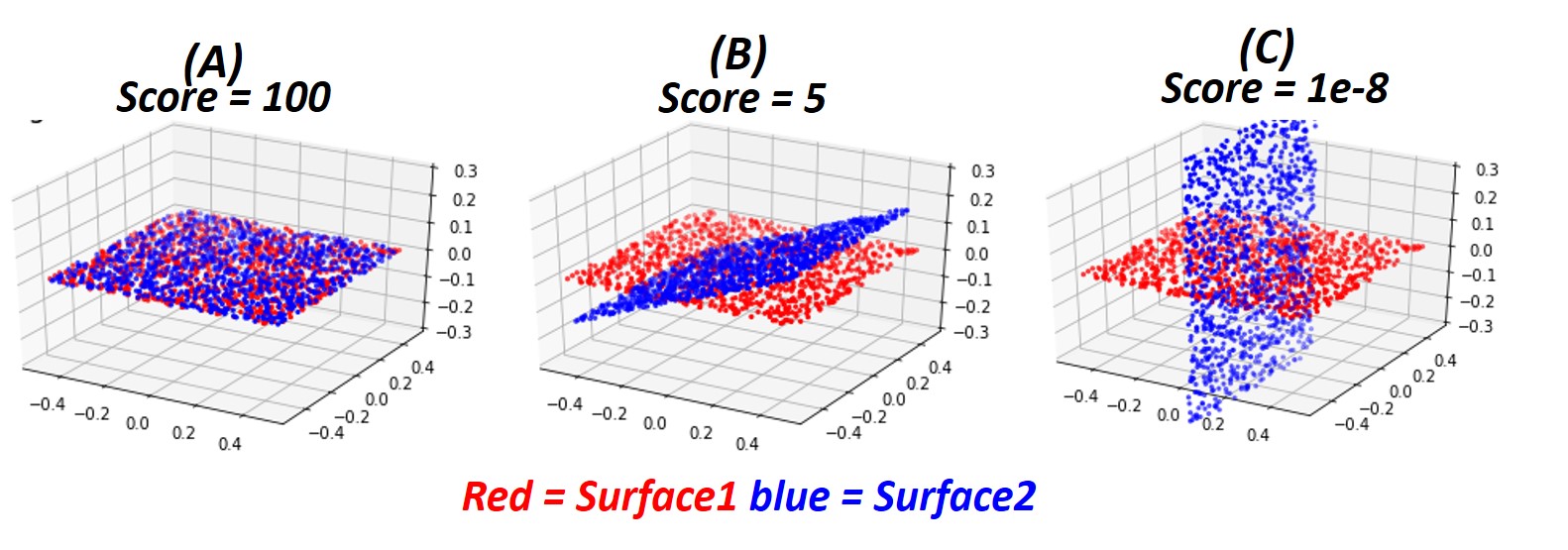} 
\end{center}
   \caption{\textit{ActivationNet} predictions for rotation of 3-D surfaces}
\label{fig:fig6}
\end{figure}

In the next experiment, we rotate the point clouds of 3-D surfaces with respect to one another and validate the performance of our model. It may be observed from Figure \ref{fig:fig6}, that the surfaces, initially, have the same size and are exactly on top of each other. The prediction of \textit{ActivationNet} is a contact score of $100$ because the angle between the surfaces is $0^o$ and there is $100$\% overlap between the surfaces. As the angle between the surfaces is increased, the overlapping regions between the surfaces changes from a surface contact to a line contact, and contact quality score prediction drops. The predicted score is the smallest when the surfaces are perpendicular to one another.

\subsection{Scaling of point clouds of 3-D surfaces}

\begin{figure}[h]
\begin{center}
\includegraphics[width=1.0\linewidth]{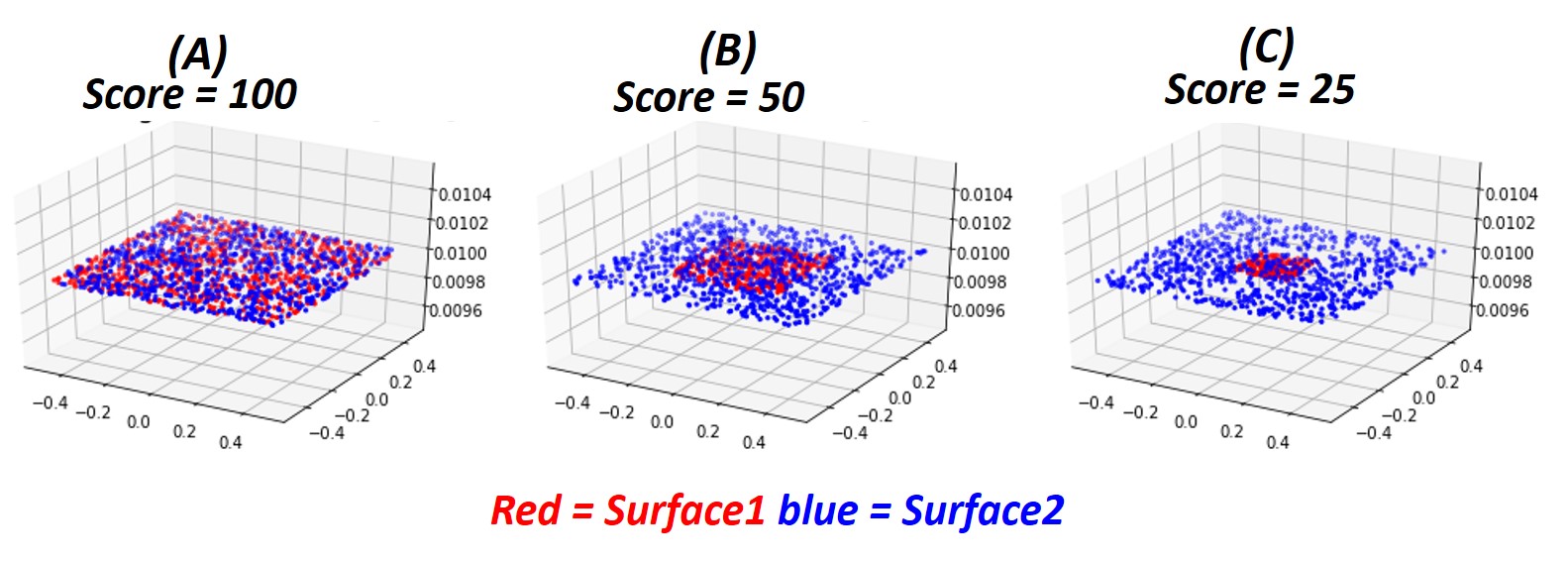} 
\end{center}
   \caption{\textit{ActivationNet} predictions for scaling of 3-D surfaces}
\label{fig:fig7}
\end{figure}

In this experiment, we scale the point clouds of 3-D surfaces with respect to one another and validate the performance of our model. It may be observed from Figure \ref{fig:fig7}, that the surfaces initially have the same size and there is $100$\% overlap between them. Subsequently, the size of surface $1$ is kept the same but the size of surface $2$ is reduced by a factor of $2$. When the surfaces have the same size, the contact quality score is $100$ due to the strong overlap. Moreover, the contact quality score reduces by the same factor as the size of the second surface. As the size of surface $2$ is made smaller, the overlapping region between the surfaces reduces and the contact quality score prediction drops by the same factor. The predictions from \textit{ActivationNet} agree well with the expectation.

\subsection{Testing on unseen engineering geometries}

It is evident from the previous experiments that the \textit{ActivationNet} is learning the true representation of interactions between arbitrary 3-D surfaces. Next, we test the model on interacting surfaces generated from $2$ engineering geometries. The engineering geometries used for testing are shown in Figure \ref{fig:fig4} and the surfaces on these geometries are not seen by our model during training. The 3-D interacting surfaces are generated using a contact detection algorithm and a coarse triangular mesh is generated on all the surfaces. A sampling technique is implemented to sample $10$ points on each triangular element of the coarse mesh. $14$ interacting surfaces are extracted from testing geometry $1$, while, $10$ interacting surfaces are extracted from testing geometry $2$. The contact quality predictions for these surfaces are carried out using \textit{ActivationNet} and PointNet \cite{qi2017pointnet}. The samples used for testing are annotated by engineering simulation experts, such that the contact quality score falls in a specified range. The goal for both the methods is to predict contact quality scores, which fall within the expected range.

\begin{figure}[h]
\begin{center}
\includegraphics[width=1.0\linewidth]{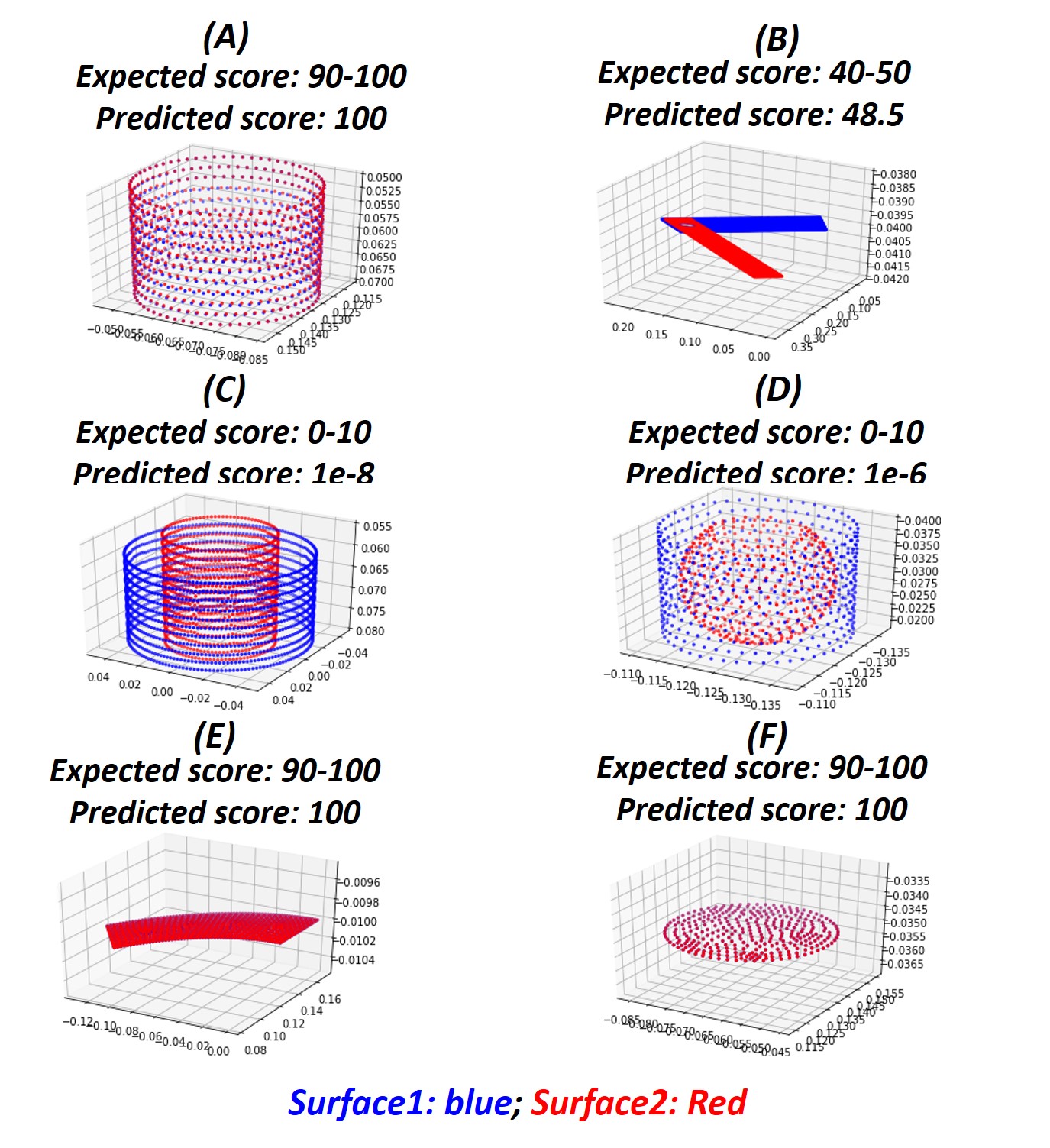} 
\end{center}
   \caption{\textit{ActivationNet} predictions on point clouds of testing geometry $1$}
\label{fig:fig8}
\end{figure}

Figure \ref{fig:fig8} shows selected results obtained on the interacting surfaces obtained from testing geometry $1$. It may be observed that for all cases, the contact quality score prediction of our model fall within the excepted range of the score assigned by expert design modelers. For test samples $A$, $E$ and $F$, the expected score ranges between $90$ and $100$, because the interacting surfaces have the same size and are positioned with a $100$\% overlapping between them. Our model is able to extract features to understand this and provide a suitable score. On the other hand, test samples $C$ and $D$ have concentric surfaces with no contact between them. Our model predicts a very low score of $1e^{-8}$ and $1e{-6}$, respectively, which is also expected for these samples. For test samples, $B$, the surfaces are at a $45^o$ angle with respect to one another and there is some overlap near the top of the surface. As a result, out model predicts a score of $48.5$ for this sample. All in all, the predictions obtained from the \textit{ActivationNet} are reasonable and reliable for testing geometry $1$ and agree well with the expected range specified by the experts modelers.   

\begin{figure}[h]
\begin{center}
\includegraphics[width=1.0\linewidth]{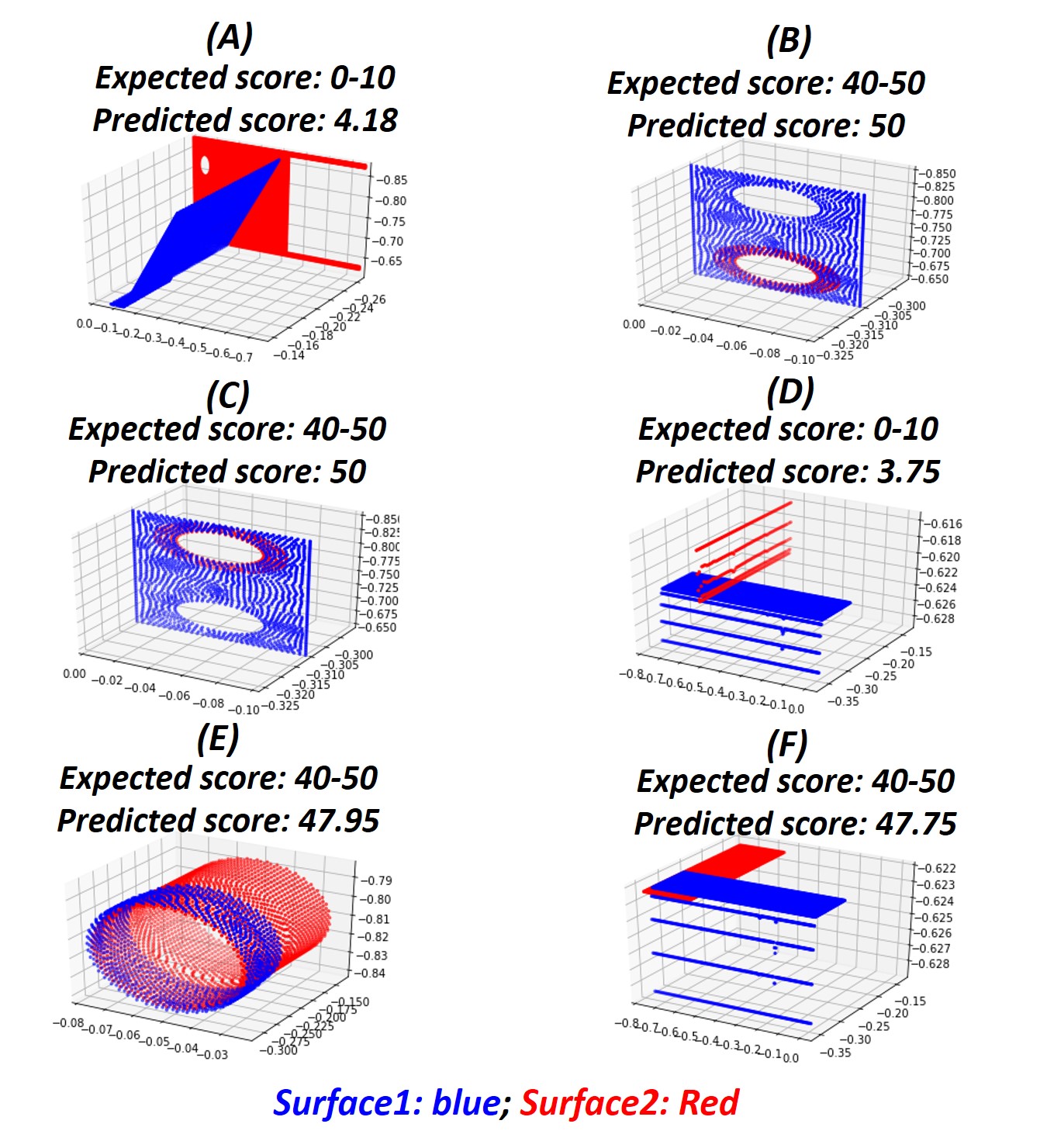} 
\end{center}
   \caption{\textit{ActivationNet} predictions on point clouds of testing geometry $2$}
\label{fig:fig9}
\end{figure}

Next, we observe some of the results obtained on the interacting surfaces obtained from testing geometry $2$. The results may be observed in Figure \ref{fig:fig9}. It may be observed that for all cases, the contact quality score prediction of our model falls within the excepted range provided by the experts. For test samples $A$ and $D$, the expected score ranges between $0$ and $10$, because the contact between the surfaces is a line contact. The features extracted by our model is able to understand this and provide a suitable score. On the other hand, test samples $F$ has a relatively bigger overlapping region and hence our model predicts a score of about $50$, which was also expected for this sample. For test samples, $B$ and $C$, the surfaces are very close to one another and there is a good amount of overlap. There is $100$\% overlap from the perspective of the surface represented by red but lesser from the perspective of the blue surface. Our model is able to capture these representations and predicts a score of $50$ for both these test samples. Moreover, sample $B$ is a flipped version of sample $C$, but the nature of contact between them is very similar. Our model can capture this and is able to make accurate and reasonable predictions. Similarly, in sample $E$, where the blue concentric surfaces has a $100$\% overlap with the red surface but the red surface is larger and the percentage of overlap maybe smaller. The proximity between the surfaces is small. Hence, the predicted contact quality score is $47.95$. Overall, the predictions obtained from the \textit{ActivationNet} are reasonable and reliable for both testing geometry's $1$ and $2$ and agree well with the expected range specified by the experts.  

Finally, we compare the results of \textit{ActivationNet} with those obtained from PointNet \cite{qi2017pointnet}. The PointNet is set up such that the point clouds from the two interacting surfaces are labeled with surface identifiers, so as to distinguish between the surfaces. In our experience, from the context of learning interactions between surfaces, it is very important to associate identifiers with the surfaces to achieve better performance with PointNet. As mentioned earlier, the network architecture for PointNet is designed to solve a $3-$class classification problem. The description of class definitions is provided in section \ref{datagen}. The contact quality score is evaluated using Equation \ref{eq:eq1} on $24$ interacting surfaces obtained from the $2$ test geometries using both PointNet and \textit{ActivationNet}. The comparisons are shown in Tables \ref{table:table1} and \ref{table:table2}. It may be observed that the \textit{ActivationNet} is effective and accurate at predicting interactions between contacting 3-D surfaces.

\begin{table}
\begin{center}
\begin{tabular}{|c|c|c|c|}
\hline
Surface Id & \textit{ActivationNet} & PointNet & Expected \\
\hline\hline
1 & 49.99 & 4.79 & 40-50 \\
2 & 100 & 84.69 & 90-100 \\
3 & 1e-6 & 0.11 & 0-10\\
4 & 1e-6 & 0.18 & 0-10\\
5 & 1e-8 & 0.02 & 0-10\\
6 & 49.004 & 6.99 & 40-50\\
7 & 49.004 & 6.17 & 40-50\\
8 & 100 & 78.18 & 90-100\\
9 & 100 & 78.83 & 90-100\\
10 & 48.5 & 0.48 & 50-60\\
11 & 100 & 58.46 & 90-100\\
12 & 100 & 99.34 & 90-100\\
13 & 100 & 53.91 & 90-100\\
14 & 100 & 99.71 & 90-100\\
\hline
\end{tabular}
\end{center}
\caption{Comparison of contact quality scores between \textit{ActivationNet} and PointNet on different surfaces of test geometry $1$}
\label{table:table1}
\end{table}

\begin{table}
\begin{center}
\begin{tabular}{|c|c|c|c|}
\hline
Surface Id & \textit{ActivationNet} & PointNet & Expected \\
\hline\hline
1 & 4.18 & 49.82 & 0-10 \\
2 & 7.93 & 49.32 & 0-10 \\
3 & 0.002 & 65.96 & 0-10\\
4 & 47.95 & 49.92 & 40-50\\
5 & 50.0 & 7.71 & 40-50\\
6 & 47.75 & 15.4 & 40-50\\
7 & 3.75 & 39.31 & 0-10\\
8 & 50 & 49.84 & 40-50\\
9 & 50 & 49.74 & 40-50\\
10 & 16.74 & 65.82 & 0-10\\
\hline
\end{tabular}
\end{center}
\caption{Comparison of contact quality scores between \textit{ActivationNet} and PointNet on different surfaces of test geometry $2$}
\label{table:table2}
\end{table}

\section{Conclusion} \label{conclusion}

In this paper, we introduce a machine learning algorithm, \textit{ActivationNet}, to extract features from point-based representations of interacting 3-D surfaces and provides a score for contact quality. Many applications in engineering simulation require information of contact quality to accurately solve physics. The traditional algorithms that can process points have not been designed to capture such interactions. Our model is geared toward such applications and provides an opportunity to effectively predict quality of contacts, thereby eliminating human intervention and saving computational time and resources in engineering simulations.

\textit{ActivationNet} processes point-based representations of interacting surfaces and generates activation states using a multi-dimensional binning approach. The activation states are generated at multiple resolutions of bins. The multi-resolution activation bins are inputs to our neural network which is a classification based algorithm. The class probabilities are used to predict a contact quality score using a heuristic model.

Finally, the \textit{ActivationNet} has been tested over several test cases. The first set of experiments are designed to prove that the \textit{ActivationNet} is learning true representations of interacting surfaces and not just overfitting the training data. This is followed by demonstrations on interacting surfaces generated from actual engineering geometries, which were not seen during training. The results of this experiment are compared with the well-known PointNet algorithm. The results from our algorithm are shown to agree well with the expected results for various interacting surfaces with arbitrary shapes and size, that were part of the testing geometries.

{\small
\bibliographystyle{elsarticle-num}
\bibliography{references}
}

\end{document}